\documentclass[lettersize,journal]{IEEEtran}
\usepackage{amsmath,amsfonts}
\usepackage{algorithm}
\usepackage{algpseudocode}
\usepackage{array}
\usepackage{textcomp}
\usepackage{stfloats}
\usepackage{url}
\usepackage{verbatim}
\usepackage{graphicx}
\usepackage{cite}
\usepackage{subfigure}

\usepackage{graphics} 
\usepackage{epsfig} 
\usepackage{mathptmx} 
\usepackage{times} 
\usepackage{amsmath} 
\usepackage{amssymb}  
\usepackage{multirow}
\usepackage{adjustbox}
\newcommand{\p}{\Delta \mathbf{p}}
\newcommand{\R}{\Delta \mathbf{R}}
\newcommand{\bp}{\overline{\Delta \mathbf{p}}}
\newcommand{\bR}{\overline{\Delta \mathbf{R}}}

\usepackage{color}
\usepackage{bm}

\usepackage[switch]{lineno}

\begin{document}

\title{Design and Evaluation of an Invariant Extended Kalman Filter for Trunk Motion Estimation with Sensor Misalignment}

\author{Zenan Zhu$^{1,*}$, Seyed Mostafa Rezayat Sorkhabadi$^{2,*}$, Yan Gu$^{3,\dagger}$, Wenlong Zhang$^{2}$
\thanks{This work was supported by the National Science Foundation under Grants IIS-1756031, IIS-1955979, CMMI-1944833, and CMMI-2046562.}
\thanks{$^{1}$Z. Zhu is with the College of Engineering,
        University of Massachusetts Lowell, Lowell, MA 01854, USA
        {\tt\small zenan\_zhu@student.uml.edu.}}%
\thanks{$^{3}$Y. Gu is with the School of Mechanical Engineering,
        Purdue University, West Lafayette, IN 47907, USA.
        This work was partly conducted while Y. Gu was with the University of Massachusetts Lowell.
        {\tt\small yan.gu.purdue@gmail.com.}}%
\thanks{$^{2}$M. Rezayat and W. Zhang are with the School of Manufacturing Systems and Networks, Ira A. Fulton Schools of Engineering, Arizona State University
        Mesa, AZ 85212, USA
        {\tt\small \{sm.rs, wenlong.zhang\}@asu.edu}.}%
\thanks{$*$ These two authors have equal contributions.}
\thanks{$\dagger$ Address all correspondence to this author.}
}

\markboth{Journal of \LaTeX\ Class Files,~Vol.~14, No.~8, August~2021}%
{Shell \MakeLowercase{\textit{et al.}}: A Sample Article Using IEEEtran.cls for IEEE Journals}

\maketitle

\begin{abstract}
Understanding human motion is of critical importance for health monitoring and control of assistive robots, yet many human kinematic variables cannot be directly or accurately measured by wearable sensors. In recent years, invariant extended Kalman filtering (InEKF) has shown a great potential in nonlinear state estimation, but its applications to human poses new challenges, including imperfect placement of wearable sensors and inaccurate measurement models. To address these challenges, this paper proposes an augmented InEKF design which considers the misalignment of the inertial sensor at the trunk as part of the states and preserves the group affine property for the process model. Personalized lower-extremity forward kinematic models are built and employed as the measurement model for the augmented InEKF. Observability analysis for the new InEKF design is presented. The filter is evaluated with three subjects in squatting, rolling-foot walking, and ladder-climbing motions. Experimental results validate the superior performance of the proposed InEKF over the state-of-the-art InEKF. Improved accuracy and faster convergence in estimating the velocity and orientation of human, in all three motions, are achieved despite the significant initial estimation errors and the uncertainties associated with the forward kinematic measurement model.  
\end{abstract}


\section{Introduction} 
\label{sec: introduction}

Wearable robots have gained growing interests over the past decades as they demonstrated great potentials in facilitating neurorehabiltiation, assisting in daily activities, and reducing work-related injuries~\cite{young2016state,nordin2014assessment,yumbla2021human}. Wearable robots have been designed with different actuation mechanisms (e.g., cable-driven  and pneumatic-driven) and materials (e.g., carbon fibers and fabrics), and they have been applied to various human joints. Since wearable robots physically interact with humans, it is critical to develop control systems that can understand the human's intent and physical states to adaptively exert an appropriate amount of assistance. To this end, various human-centered controller designs have been applied in wearable robots, including adaptive impedance control~\cite{chinimilli2019automatic}, phase-based control~\cite{martinez2018velocity}, reinforcement learning~\cite{luo2021reinforcement}, to name a few.

All the aforementioned robot controllers rely on wearable sensors and estimation algorithms to understand human's intent and physical states. Many existing controllers (e.g., finite-state machines) for wearable robots rely on classification of human motion into finite number of states, such as gait phases~\cite{del2014hybrid} and activity types~\cite{cheng2021real}. However, this method will limit the humans to pre-defined activities and it can suffer from misclassification if the user behaves differently from the training data. On the other side of the spectrum, continuous human states, such as positions and orientations, present a promising direction for controller design since they contain more information about the user for fine adjustment of the robot assistance. A challenge for using continuous human states is that many of them cannot be directly measured, or the sensor measurement is too noisy to be used for robot control. This is particularly important for lower-extremity wearable robots, where the center of mass is critical for ensuring postural stability but is not directly measurable~\cite{huang2021estimating}.

Various approaches have been proposed for the estimation of continuous human movement state.
Earlier methods produce accurate, real-time estimation of stance-foot locations during human walking, by fusing the zero toe velocity during stance phase and the reading of inertial measurement units (IMUs) attached to a subject's shoes~\cite{ojeda2007personal}.
To monitor the movement of the body (e.g., pelvis) during walking and stair climbing, the stance leg's kinematics can be used to compute the movement under the assumptions that the stance feet is stationary and that the contact detection is sufficiently accurate~\cite{yuan20133}. 
This method is computationally efficient for real-time estimation, and its accuracy has been improved by explicitly handling the joint angle reading noise through Kalman filtering~\cite{yuan2014localization}.
To further enhance the error convergence, extended Kalman filtering (EKF) has been introduced to fuse the forward kinematics of a subject's segments with the data returned by an IMU attached to the body during bicycle riding~\cite{zhang2015whole}. 
Still, the standard EKF methodology relies on system linearization whose accuracy depends on the estimation error, and thus its performance under large initial errors may not be satisfactory.
To this end, the invariant EKF (InEKF)~\cite{barrau2016invariant} has been created to ensure accurate, efficient estimation under large errors, by exploiting the accurate linearization of systems that meet the group affine condition and invariant observation form.

Although the existing InEKF methods have achieved rapid convergence and accurate estimates of legged robots under large estimation errors~\cite{hartley2020contact,gao2021invariant,teng2021legged}, InEKF design for human movement state estimation poses complex challenges. 
One challenge is the inaccurate or unknown sensor placement relative to the subject's segments during subject movement.
The relative inertial sensor placement is assumed to be accurately known in the existing fusion of inertial odometry and limb kinematics of legged robots~\cite{hartley2020contact,gao2021invariant,teng2021legged}.
Yet, this assumption may not be realistic for human movement estimation because it may require frequent calibration during relatively dynamic movement (e.g., squatting, walking, and ladder climbing) due to shift of sensors on the skin or garment. There is a rich body of work in fault detection, but most of them focus on detecting faults in the system rather than obtaining accurate state estimation with such faults~\cite{cheng2021asynchronous}.
Thus, an appropriate filter design should address the sensor placement imperfection.
Its performance also needs to be examined during various types of common daily mobility tasks so that accurate, real-time state estimates could be used to inform the control of wearable devices for different human movement activities.

This paper introduces an InEKF design to address sensor misalignment, which is a very common problem in human and robot locomotion, so that sensor placement offset can be estimated and corrected to ensure accurate state estimation. The intended application is to use lower-extremity joint angles and forward kinematic models to augment the trunk velocity and orientation estimation by correcting misalignment of trunk sensors during the stance phase.
A preliminary version of this work was presented in~\cite{zhu2021acc}.
 The present paper reports the following new, substantial contributions.
 (i) A forward kinematic model is developed in this work and used as the measurement model for the InEKF. The new measurement model is less accurate but much more practical since it relies on joint angles, compared to the 3-D vectors used in~\cite{zhu2021acc}.
 (ii) An observability analysis is presented for the new InEKF design and validated using experimental results with large initial errors.
 (iii) Two new activities, namely ladder climbing and rolling-foot walking, are evaluated in human testing along with squat motion in \cite{zhu2021acc}. The new results extend the InEKF to more dynamic locomotion tasks with new challenges in the measurement model introduced by contact-foot switching. 
 (iv) The proposed filter is evaluated with three participants in three tasks, and results are compared with the existing InEKF to draw insights on how to extend InEKF to human locomotion estimation.

The paper is structured as follows.
Section~\ref{sec: modeling} introduces the state representation and the process and measurement models of the proposed InEKF.
Section~\ref{sec: InEKF} presents the propagation and update steps of the filter along with the observability analysis.
Section~\ref{sec: results} reports and discusses the experiment validation results.
Section~\ref{sec: conclusion} provides the conclusion remarks.

\section{System Modeling}
\label{sec: modeling}

\subsection{Preliminaries}

\begin{figure}[t]
    \centering
    \includegraphics[width=0.9\linewidth]{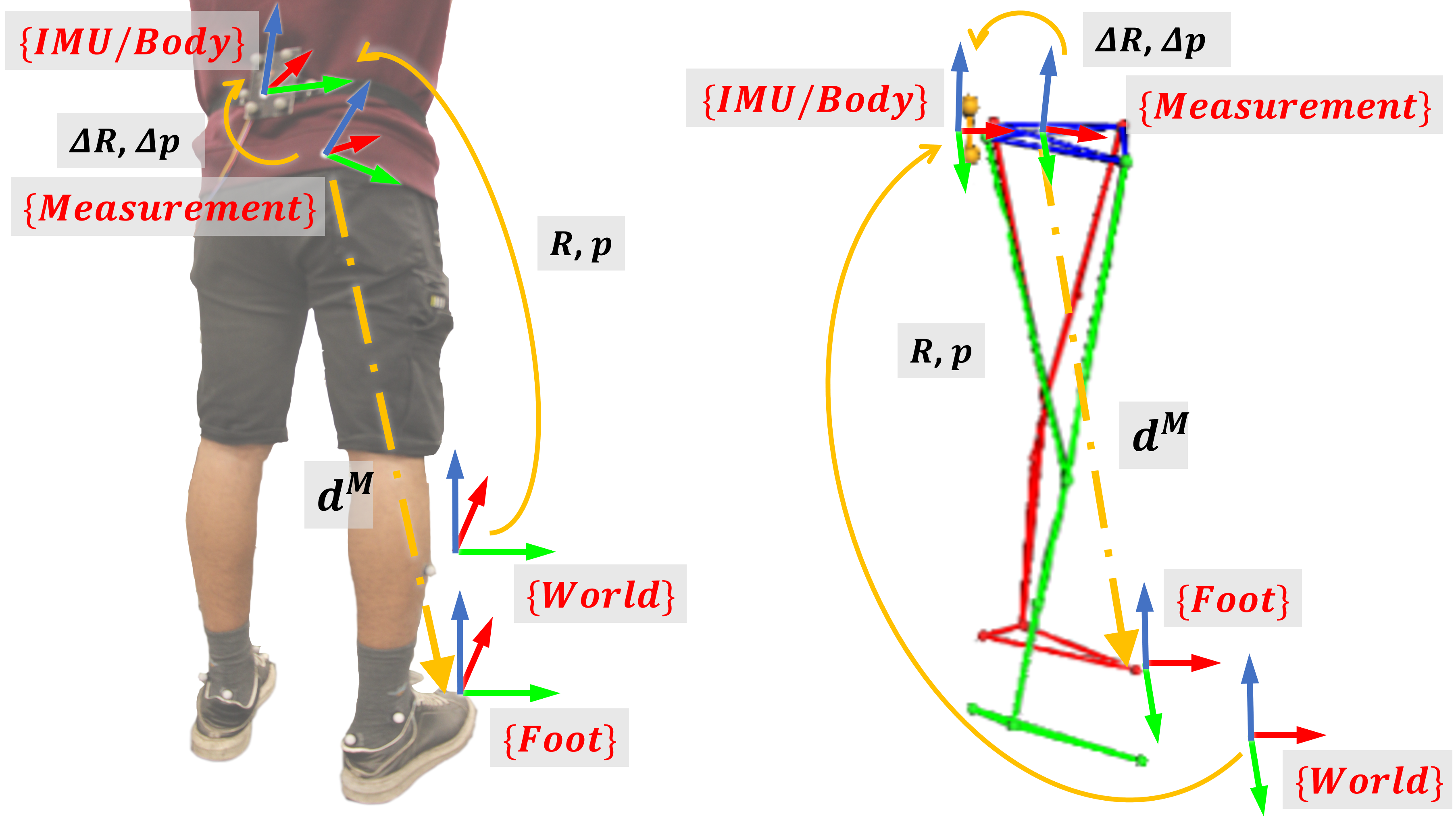}

    \caption{Coordinate frames and estimation variables used in the proposed filter design.
    They are illustrated on the real human subject (left) and the subject's lower extremity skeleton in the motion capture software (right). $\mathbf{R}$ and $\mathbf{p}$ represent the orientation and position of the IMU (body) in the world frame, and $\mathbf{d}^M$ represents the vector from the measurement frame to the foot frame, expressed in the measurement frame. $\boldsymbol{\Delta}\mathbf{p}$ and $\boldsymbol{\Delta}\mathbf{R}$ represent the position and orientation offset of the IMU frame with respect to the measurement frame, expressed in the measurement frame. }
    \label{fig: coordinates represents}

\end{figure}

A matrix Lie group ${G}$ is a subset of $ n \times n$ invertible matrices possessing the following properties: $\forall \mathbf{X} \in G, \mathbf X^{-1} \in G$; $\forall \mathbf X_1, \mathbf X_2 \in G,   \mathbf X_1 \mathbf X_2 \in G$; and $\mathbf{I} \in G$ with $\mathbf{I}$ the identity element.
The associated Lie algebra $\boldsymbol{\mathfrak{g}}$ with a dimension of $\mbox{dim}$ ${\boldsymbol{\mathfrak{g}}}$ is the tangent space defined at $\mathbf{E}$, and is a set of $ n \times n$ square matrices.
Any vector $\boldsymbol{\zeta}\in \mathbb{R}^{\text{dim}\, {\boldsymbol{\mathfrak{g}}}} $ can be mapped onto the Lie algebra through the linear operator $(.)^{\wedge}:\mathbb{R}^{\text{dim}\, {\boldsymbol{\mathfrak{g}}}} \rightarrow \mathfrak{g}$.
The exponential map, $\text{exp}:\mathbb{R}^{\text{dim}\, {\boldsymbol{\mathfrak{g}}}}\rightarrow {G}$, is defined as: $\text{exp}(\boldsymbol{\zeta}) \triangleq \text{expm}(\boldsymbol{\zeta}^\wedge),$
where \text{expm} is the usual matrix exponential. 
The inverse operator of $ (\cdot)^\wedge$ is denoted as $(\cdot)^\vee:\boldsymbol{\mathfrak{g}}\rightarrow 
\mathbb{R}^{\text{dim}\, {\boldsymbol{\mathfrak{g}}}}$.
The adjoint matrix $\mathbf{Ad}_{\mathbf{X}}$ at $\mathbf{X}$ for any vector $\boldsymbol{\zeta} \in
\mathbb{R}^{\text{dim}\, {\boldsymbol{\mathfrak{g}}}}$ 
is defined as
$\mathbf{Ad}_{\mathbf{X}}\boldsymbol{\zeta}=(\mathbf{X}\boldsymbol{\zeta}^\wedge\mathbf{X}^{-1})^\vee$,
which is the linear mapping from the local tangent space (defined at $\mathbf{X}$) to the Lie algebra.
A more detailed introduction to matrix Lie groups is given in~\cite{sola2018micro}.

\subsection{Sensors considered}

The sensors considered in the proposed filter design include:
a) an inertial measurement unit (IMU) attached to the subject's body (e.g., lower back near the pelvis), which measures the linear acceleration $\mathbf{a} \in \mathbb{R}^3$ and angular velocity $\boldsymbol{\omega} \in SO(3)$ of the IMU/body with respect to the IMU frame, and 
b) a motion capture system (e.g., IMU sensors or markers for motion capture cameras) that measures the lower-extremity joint angles of the subject.

The particular issue this study focuses on addressing is the placement inaccuracy of the IMU that measures the subject's body movement.
As reviewed in Sec.~\ref{sec: introduction}, to achieve accurate state estimation during mobility tasks (e.g., walking and ladder climbing) despite sensor noises, an effective technique is to fuse the data returned by IMU and the subject's limb kinematics obtained based on motion capture data. 
For effective sensor fusion, the IMU should be static relative to a reference segment along the limb kinematics chain, and the relative pose needs to be accurately known~\cite{hartley2020contact}.
Yet, the exact relative sensor placement is usually unknown without calibration.

To explicitly treat the sensor placement offset, the proposed filter design involves the body/IMU frame, which is fixed to the IMU, and the measurement frame, which is fixed to the reference segment of the kinematic chains sensed by the motion capture system (see Fig.~\ref{fig: coordinates represents}).
As the filter aims to estimate subject movement in the environment, the world frame is considered.
Also, the foot frame, which is rigidly attached to the stance foot, is used because the foot defines the far end of the kinematics chains of interest.

\subsection{State representation}

To estimate body movement in the world, one state variable of interest to this study is the subject's body position, denoted as ${}^W\mathbf{p}^{WB} \in \mathbb{R}^3$.
Here, the superscript on the left-hand side represents the coordinate frame in which the variable is expressed, with $W$, $B$, $M$, and $F$ denoting the world, body/IMU, measurement, and foot frames.
The superscript $WB$ on the right-hand side indicates the position vector points from the origin of the world frame to that of the body/IMU frame.
The state variables also include the body velocity, ${}^W\mathbf{v}^{WB} \in \mathbb{R}^3$, which is the time derivative of ${}^W\mathbf{p}^{WB}$, as well as the body orientation ${}^W\mathbf{R}^{B} \in SO(3)$, which is the orientation of the body/IMU frame with respect to the world frame.
In addition, to explicitly treat the IMU placement offset (i.e., the relative position and orientation of the IMU/body frame with respect to the measurement frame), they are also chosen as state variables and denoted as ${}^M\mathbf{p}^{MB} \in \mathbb{R}^3$ and ${}^M\mathbf{R}^{B} \in SO(3)$.
For notational brevity, the rest of the paper uses $\mathbf{p}$, $\mathbf{v}$, $\mathbf{R}$, $\boldsymbol{\Delta}\mathbf{p}$, and $\boldsymbol{\Delta}\mathbf{R}$ to respectively denote ${}^W\mathbf{p}^{WB}$, ${}^W\mathbf{v}^{WB}$, ${}^W\mathbf{R}^{B}$, ${}^M\mathbf{p}^{MB}$, and ${}^M\mathbf{R}^{B}$.

We express the state variables on the matrix Lie group $G$ as prescribed by the methodology of InEKF~\cite{barrau2016invariant}:
\begin{equation}
    \mathbf{X}=
    \begin{bmatrix}
    \mathbf{R} & \mathbf{v} & \mathbf{p} & \mathbf{0}_{3,3} & \mathbf{0}_{3,1}\\
    \mathbf{0}_{1,3} & 1 & 0 & \mathbf{0}_{1,3} & \mathbf{0}_{3,1}\\
    \mathbf{0}_{1,3} & 0 & 1 & \mathbf{0}_{1,3} & \mathbf{0}_{3,1}\\
    \mathbf{0}_{3,1} & \mathbf{0}_{3,1} & \mathbf{0}_{3,1} & \R & \p\\
    \mathbf{0}_{1,3} & 0 & 0 & \mathbf{0}_{1,3} & 1
    \end{bmatrix}
    \in G.
\label{eqn: X}
\end{equation}
Here, the matrix Lie group $G$ is a combination of double direct spatial isometries $SE_2(3)$ \cite{barrau2016invariant} and a special Euclidean group $SE(3)$,
and $\mathbf{0}_{n,m}$ represents an $n\times m$ zero matrix.
It can be proved that $G$ is a valid matrix Lie group.

\subsection{Process model}

This subsection introduces the proposed process model in Euclidean space as well as on the matrix Lie group ${G}$.

\subsubsection{IMU motion dynamics}

Given its accuracy and simplicity~\cite{hartley2020contact}, the IMU motion dynamics is used to build a process model with the noisy IMU readings $\tilde{\mathbf{a}}$ and $\tilde{\boldsymbol{\omega}}$ serving as its input.
Corrupted by white Gaussian zero-mean noise $\mathbf{w}_a \in \mathbb{R}^3$ and $\mathbf{w}_{\omega} \in SO(3)$, these readings are given by: $\mathbf{\tilde{a}}= \mathbf{a}+\mathbf{w}_a$ and $\boldsymbol{\tilde{\omega}}= \boldsymbol{\omega}+\mathbf{w}_{\omega}$. Note that for simplicity, the biases in the raw data returned by the accelerometer and gyroscope are not considered here.
Then, the IMU motion dynamics can be expressed as:
$
\frac{d}{dt}\mathbf{R}= \mathbf{R}(\tilde{\boldsymbol{\omega}}-\mathbf{w}_{\omega})_\times$,
$\frac{d}{dt}\mathbf{p}=\mathbf{v}$,
and
$\frac{d}{dt}\mathbf{v}=\mathbf{R}(\tilde{\mathbf{a}}-\mathbf{w}_a)+\mathbf{g}$,
where $(.)_\times$ is a skew-symmetric matrix and $\mathbf{g}$ is the gravitational acceleration. 

\subsubsection{IMU placement error dynamics}

Given that the IMU, if appropriately attached, usually does not shift quickly on the subject, we choose to model the dynamics of the IMU placement offsets $\p$ and $\R$ as slowly time-varying, which is given by:
$\frac{d}{dt}{\p}=\mathbf{w}_{\Delta p}$
and
$\frac{d}{dt}{\R}={\R}(\mathbf{w}_{\Delta R})_\times$,
where $\mathbf{w}_{\Delta p} \in \mathbb{R}^3$ and $\mathbf{w}_{\Delta R} \in SO(3)$ are white Gaussian noise with zero mean. 

\subsubsection{Process model on $G$}
At time $t$, these process models can be compactly expressed on the matrix Lie group ${G}$ as:
 \begin{equation}
    \begin{aligned}
    \frac{d}{dt}\mathbf{X}_t
    =&\begin{bmatrix}
\mathbf{R}_t(\tilde{\boldsymbol{\omega}}_t)_\times & \mathbf{R}_t \tilde{\mathbf{a}}_t+\mathbf{g} & \mathbf{v}_t & \mathbf{0}_{3,3} & \mathbf{0}_{3,1}\\
\mathbf{0}_{1,3} & 0 & 0 & \mathbf{0}_{1,3} & \mathbf{0}_{3,1}\\
\mathbf{0}_{1,3} & 0 & 0 & \mathbf{0}_{1,3} & \mathbf{0}_{3,1}\\
\mathbf{0}_{3,3} & \mathbf{0}_{3,1} & \mathbf{0}_{3,1} & \mathbf{0}_{3,3} & \mathbf{0}_{3,1}\\
\mathbf{0}_{1,3} & 0 & 0 & \mathbf{0}_{1,3} & 0
\end{bmatrix}\\
&-\mathbf{X}_t
\begin{bmatrix}
(\mathbf{w}_{\omega_t})_\times & \mathbf{w}_{a_t} & \mathbf{0}_{3,1} & \mathbf{0}_{3,3} & \mathbf{0}_{3,1}\\
\mathbf{0}_{1,3} & 0 & 0 & \mathbf{0}_{1,3} & \mathbf{0}_{3,1}\\
\mathbf{0}_{1,3} & 0 & 0 & \mathbf{0}_{1,3} & \mathbf{0}_{3,1}\\
\mathbf{0}_{3,3} & \mathbf{0}_{3,3} & \mathbf{0}_{3,3} & (\mathbf{w}_{\Delta R_t})_\times & \mathbf{w}_{\Delta p_t}\\
\mathbf{0}_{1,3} & 0 & 0 & \mathbf{0}_{1,3} & 0
\end{bmatrix}\\
\triangleq & {f}_{u_t}(\mathbf{X}_t)-\mathbf{X}_t \mathbf{w}_t ^\wedge, \label{Dynamics}
    \end{aligned}
\end{equation}
where the vector $\mathbf{w}_t$ is defined in the IMU/body frame as $\mathbf{w}_t \triangleq \text{vec}(\mathbf{w}_{\omega_t},\mathbf{w}_{a_t} ,\mathbf{0}_{3,1}, \mathbf{w}_{\Delta R_t} , \mathbf{w}_{\Delta p_t}) $,
and the vector $\mathbf{u}_t$ is the input to the process model defined as $\mathbf{u}_t \triangleq \text{vec}(\tilde{\boldsymbol{\omega}}_t,\tilde{\mathbf{a}}_t)$.

\noindent \textbf{Proposition 1 (Group affine system)}
The deterministic portion of the process model in ~\eqref{Dynamics} is group affine; that is, the deterministic dynamics ${f}_{u_t}(.)$ meets the following group affine condition for right-invariant cases \cite{barrau2016invariant}:

\begin{equation}
{f}_{u_t}(\mathbf{X}_1 \mathbf{X}_2)
={f}_{u_t}(\mathbf{X}_1)\mathbf{X}_2 + \mathbf{X}_1 {f}_{u_t}(\mathbf{X}_2) -\mathbf{X}_1 {f}_{u_t}(\mathbf{I})\mathbf{X}_2.
\label{eqn: group affine}
\end{equation}

\noindent \textbf{Proof.}
By the definition of $f_{u_t}$ in \eqref{Dynamics}, we obtain
\begin{equation*}
\begin{gathered}
    f_{u_t}(\mathbf{X}_1)=\begin{bmatrix}
\mathbf{R}_1(\tilde{\boldsymbol{\omega}})_\times & \mathbf{R}_1 \tilde{\mathbf{a}}+\mathbf{g} & \mathbf{v}_1 & \mathbf{0}_{3,4}\\
\mathbf{0}_{6,3} & \mathbf 0_{6,1} & \mathbf 0_{6,1} & \mathbf{0}_{6,4}\\
\end{bmatrix}, \\
f_{u_t}(\mathbf{X}_1 \mathbf{X}_2)=\begin{bmatrix}
\mathbf{R}_1\mathbf{R}_2(\tilde{\boldsymbol{\omega}})_\times & \mathbf{R}_1\mathbf{R}_2 \tilde{\mathbf{a}}+\mathbf{g} &\mathbf{R}_1 \mathbf{v}_2+ \mathbf{v}_1 & \mathbf{0}_{3,4}\\
\mathbf{0}_{6,3} & \mathbf 0_{6,1} & \mathbf 0_{6,1} & \mathbf{0}_{6,4}\\
\end{bmatrix},~\mbox{and}~
\\
f_{u_t}(\mathbf{I})=\begin{bmatrix}
(\tilde{\boldsymbol{\omega}})_\times & \tilde{\mathbf{a}}+\mathbf{g} & \mathbf{0}_{3,1} & \mathbf{0}_{3,4}\\
\mathbf{0}_{6,3} & \mathbf 0_{6,1} & \mathbf 0_{6,1} & \mathbf{0}_{6,4}\\
\end{bmatrix}
\end{gathered}
\end{equation*}
for any $\mathbf X_1, \mathbf X_2  \in G$.
Here $\mathbf{R}_i$ and $\mathbf{v}_i$ ($i=1,2$) denote elements of the state $\mathbf{X}_i$.
It can be seen that these matrices satisfy the condition in~\eqref{eqn: group affine}.

\subsection{Measurement model}
\subsubsection{Forward kinematics based measurement}
During mobility tasks (e.g., squatting, walking, and stair climbing), when the stance foot has a static, secured contact with the ground, the pose of the measurement frame can be obtained through the leg kinematic chain that connects the foot and measurement frames.
Note that the joint angles along the kinematic chain are measured by the motion capture system. The kinematic chain is built based on the Vicon lower-body Plug-in-Gait model \cite{kainz2017reliability}, in which 3D joint angles (hip, knee, and ankle), along with subject lower-body segment length measurements, were used  to obtain the desired 3D vector.

Let $\boldsymbol{\alpha}_t \in \mathbb{R}^{k}$ be the joint angles of the stance leg with $k$ the number of joint angles.
Then the measured joint angle $\tilde{\boldsymbol{\alpha}}_t$ can be expressed as $\tilde{\boldsymbol{\alpha}}_t=\boldsymbol{\alpha}_t+\mathbf{w}_{\alpha_t}$ with $\mathbf{w}_{\alpha_t} \in \mathbb{R}^{k}$ the zero-mean white Gaussian noise.

Let ${}^{M}\mathbf{d}^{MF}_t$ denote the 3-D position vector pointing from the measurement to the foot frame expressed in the measurement frame. 
For brevity, we denote it as $\mathbf{d}^{M}_t$.
Let the function ${h}_{F}$ be the forward kinematics representing $\mathbf{d}^{M}_t$; that is, $\mathbf{d}^{M}_t={h}_{F}(\boldsymbol{\alpha}_t)$.

Given that $\mathbf d_t^M=(\R_t)\mathbf R_t^T (\mathbf d_t-\mathbf p_t) -\p_t$, we have $\mathbf d_t -\mathbf p_t =\mathbf R_t \R_t^T(\p_t + {h}_{F}( \boldsymbol{\alpha}_t))$.
Taking the first derivative with respect to time $t$ on both sides of this equation, we obtain: 
\begin{equation}
\begin{aligned}
    & \quad 
    \frac{d}{dt}{(\mathbf{d}_t-\mathbf{p}_t)}
    = 
    \mathbf R_t \R_t^T (\mathbf w_{\Delta p_t} +  J(\boldsymbol{\alpha}_t) (\dot{\boldsymbol{\alpha}_t}+\mathbf w_{\dot{\alpha}_t}))
    \\
    & \quad +
    \left( \mathbf R_t(\boldsymbol{\omega}_t+\mathbf{w}_{\omega_t})_\times \R_t^T 
    + \mathbf R_t(\R_t (\mathbf w_{\R_t})_\times)^T \right) ( {h}_{F}(\boldsymbol{\alpha}_t)+\p_t),
    \label{kinematics measurement equation_0}
\end{aligned}
\end{equation}
where $ J (\boldsymbol{\alpha}) \triangleq \frac{\partial {h}_{F} (\boldsymbol{\alpha})}{\partial \boldsymbol{\alpha}}$ is the forward kinematic Jacobian
and $\mathbf{w}_{\dot{\alpha}_t}$ is the joint velocity measurement noise.

Note that $\dot{\mathbf d}_t=\mathbf 0$ due to the stationary contact point.
Also, $\dot{\mathbf p}_t=\mathbf v_t$ holds.
Thus, the measurement model in \eqref{kinematics measurement equation_0} can be compactly rewritten as:
\begin{equation}
    \mathbf y= {h}(\mathbf X_t)+\mathbf n_t,
    \label{kinematics measurement equation}
\end{equation}
where
$\mathbf y = - J(\tilde{\boldsymbol{\alpha}_t})  \dot{\tilde{\boldsymbol{\alpha}_t}}$, 
${h}(\mathbf X_t) = (\R_t)\mathbf R_t^T \mathbf v_t -  (\p_t)_\times \R_t \tilde{\boldsymbol{\omega}_t} - (  h_F(\tilde{\boldsymbol{\alpha}_t}))_\times  \R_t \tilde{\boldsymbol{\omega}_t}$,
and 
$\mathbf n_t=({\p_t}+h_F(\tilde{\boldsymbol{\alpha}_t})_\times{\R_t}(\mathbf{w}_{\Delta R_t}+\mathbf{w}_{\omega_t})+\mathbf{w}_{\Delta p_t}-J(\tilde{\boldsymbol{\alpha}_t})\mathbf{w}_{\dot{\alpha_t}}$.
For simplicity, $\mathbf n_t$ is treated as white, Gaussian, zero-mean noise in this study.

\subsubsection{3-D vector based measurement model}
To compare the filtering performance under different kinematics measurements formed based on data returned by motion capture systems, we introduce a simplified measurement obtained based on the 3-D position vector between the measurement and foot frame. Both forward kinematics based and 3-D position vector based measurements return the relative position and velocity between the measurement and foot frames.
Note that the 3-D position vector based measurement will have higher accuracy than the forward kinematics measurements. 
 
Let $\mathbf{v}^{M}_t$ denotes the 3-D velocity vector pointing from the measurement to the foot frame expressed in the measurement frame, which is the time derivative of $\mathbf{d}^{M}_t$.
Then the measured 3-D velocity vector $\tilde {\mathbf{v}}^{M}_t$ is expressed as $\tilde {\mathbf{v}}^{M}_t = \mathbf{v}^{M} + \mathbf {w}_{v^M_t}$, with $\mathbf {w}_{v^M_t}$ the noise associated with the 3-D velocity vector measurement. 

Given that $\mathbf d_t^M=(\R_t)\mathbf R_t^T (\mathbf d_t-\mathbf p_t) -\p_t$, we have $\mathbf d_t -\mathbf p_t =\mathbf R_t \R_t^T(\p_t + \mathbf {d}_t^M))$.
Taking the time derivative on both sides of this equation, we obtain:
\begin{equation}
\begin{aligned}
    & \quad \frac{d}{dt}{(\mathbf{d}_t-\mathbf{p}_t)} = 
    \mathbf {R}_t \R_t^T (\mathbf {w}_{\p_t} +  (\mathbf{v}^{M} + \mathbf {w}_{v^M}) ) \\
    & \quad + \left( \mathbf {R}_t(\boldsymbol{\omega}_t)_\times \R_t^T
    + \mathbf {R}_t(\R_t (\mathbf w_{\R_t})_\times)^T \right) ( \mathbf {d}_t^M + \p_t).
\end{aligned}
\label{kinematics measurement equation 3-D vector}
\end{equation}

Assume the contact foot velocity is zero due to the stationary contact point. Then the measurement model~\eqref{kinematics measurement equation 3-D vector} becomes:
\begin{equation}
\begin{aligned}
    & \quad \mathbf y= {h}(\mathbf X_t)+\mathbf n_t \\
\end{aligned}
\label{final kinematics measurement equation 3-D vector}
\end{equation}
where
$\mathbf y = - \tilde {\mathbf{v}}^{M}$, ${h}(\mathbf X_t) = \R_t \mathbf {R}_t^T \mathbf {v}_t - (\p_t)_\times \R_t \boldsymbol{\omega}_t - (\mathbf {d}_t^M)_\times \R_t \boldsymbol{\omega}_t +\mathbf n_t$,
and the vector $\mathbf n_t$ is the lumped measurement noise term.

\noindent \textbf{Remark 1:}
The two measurement models in~\eqref{kinematics measurement equation} and \eqref{final kinematics measurement equation 3-D vector} do not satisfy the right-invariant observation form (i.e., {$\mathbf{y}=\mathbf{X}^{-1} \mathbf{b}$} for some known vector $\mathbf{b}$) as defined in the theory of InEKF~\cite{barrau2016invariant}.
This is because with the state defined in {\eqref{eqn: X}} and our measurements in {\eqref{kinematics measurement equation}} and {\eqref{final kinematics measurement equation 3-D vector}}, a vector $\mathbf{b}$ that is known and satisfies {$\mathbf{y}=\mathbf{X}^{-1} \mathbf{b}$} does not exist. 
Then, by the theory of invariant filtering (Proposition 2 in \cite{barrau2016invariant}), 
the error dynamics during the measurement update is not independent of state trajectories.
The effects of this property on filter performance are discussed in Sec.~\ref{sec: results}-D.

\section{InEKF Design}
\label{sec: InEKF}

\subsection{State propagation}
The design of the proposed InEKF relies on the right-invariant error $\boldsymbol{\eta}_t$ between the true and estimated values:

\begin{equation}
\boldsymbol{\eta}_t=\Bar{\mathbf{X}}_t \mathbf{X}_t ^{-1}\in G,
\end{equation}
where $\overline{(.)}$ denotes the estimated value of the variable $(.)$.
Based on this definition and the process model (\ref{Dynamics}), one can 
obtain:

\begin{equation}
\begin{gathered}
    \frac{d}{dt} \boldsymbol{\eta}_t ={f}_{u_t}(\boldsymbol{\eta}_t)-\boldsymbol{\eta}_t {f}_{u_t}(\mathbf{I}_d) + {Ad}_{\bar{\mathbf{X}}_t} \mathbf{w}_t^\wedge
    \triangleq {g}_{u_t}(\boldsymbol{\eta}_t)+\Bar{{ \mathbf{w}}}_t^\wedge.  \label{error_noise}
    \end{gathered}
\end{equation}

\noindent \textbf{Remark 2:}
The deterministic portion of the error dynamics in \eqref{error_noise} is independent of the state trajectories because the process model in \eqref{Dynamics} is group affine, as predicted by the theory of InEKF~\cite{barrau2016invariant}.
This property is drastically different from the standard EKF whose error dynamics depends on the state trajectories.
Furthermore, since the process model is group affine, by the theory of invariant filtering (Theorem 2 in ~\cite{barrau2016invariant}), the corresponding dynamics of the log of the invariant error $\boldsymbol{\eta}_t$ is exactly linear in the deterministic case, whose expression is derived next.

Let $\boldsymbol{\zeta}_t$ be the logarithmic error defined through
$\boldsymbol{\eta}_t=\exp(\boldsymbol{\zeta}_t)$, and denote $\boldsymbol{\zeta}_t$ as
$\boldsymbol{\zeta}_t=\text{vec}(\boldsymbol{\zeta}_{{R}_t},\boldsymbol{\zeta}_{{v}_t}, \boldsymbol{\zeta}_{{p}_t},\boldsymbol{\zeta}_{\Delta {R}_t},\boldsymbol{\zeta}_{\Delta {p}_t}) \in \mathbb{R}^{\text{dim}\, {\boldsymbol{\mathfrak{g}}}}$. 
In order to linearize this error dynamics, we use the first-order approximation  $\boldsymbol{\eta}_t=\exp(  \boldsymbol{\zeta}_t)\approx \mathbf{I} +  \boldsymbol{\zeta}_t^\wedge\label{linear_exp}$, which allows us to obtain the Jacobian $\mathbf{A}_t$ of deterministic dynamics as~\cite{barrau2016invariant}:
\begin{gather}
{g}_{u_t}(\exp( \boldsymbol{\zeta}_t))=(\mathbf{A}_t  \boldsymbol{\zeta}_t)^\wedge + \text{h.o.t.}(||  \boldsymbol{\zeta}_t||)\approx (\mathbf{A}_t  \boldsymbol{\zeta}_t)^\wedge, \label{linear_Gut} 
\end{gather}
and the Jacobian $\mathbf{A}_t$ will also define the linear dynamics of the log of the right invariant error as follows:
\begin{equation}
\frac{d}{dt}  \boldsymbol{\zeta}_t=\mathbf{A}_t  \boldsymbol{\zeta}_t + \bar{\mathbf w}_t 
    \label{linear_dif_error}.
\end{equation}
To compute $\mathbf{A}_t$, we plug in the approximated right-invariant error into (\ref{linear_Gut}) as:
\begin{equation}
    \begin{aligned}
    {g}_{u_t}(\exp( \boldsymbol{\zeta}_t)) 
    &\approx {g}_{u_t}(\mathbf{I}_d + \boldsymbol{\zeta}_t^{\wedge})
    \\
    &=  \begin{bmatrix}
    \mathbf{0}_{3,3} & (\mathbf{g})_\times \boldsymbol{\zeta}_{R_t} & \boldsymbol{\zeta}_{v_t}& \mathbf{0}_{3,4}\\
    \mathbf{0}_{1,3} & 0 & 0 & \mathbf{0}_{1,4} \\
    \mathbf{0}_{1,3} & 0 & 0 & \mathbf{0}_{1,4} \\
    \mathbf{0}_{3,3} & \mathbf{0}_{3,1} & \mathbf{0}_{3,1} & \mathbf{0}_{3,4}\\
    \mathbf{0}_{1,3} & 0 & 0 & \mathbf{0}_{1,4}
    \end{bmatrix}=\begin{bmatrix}
     \mathbf{0}_{3,1}\\ (\mathbf g)_\times \\ \boldsymbol{\zeta}_{v_t} \\ \mathbf{0}_{3,1} \\ \mathbf{0}_{3,1}
    \end{bmatrix}^\wedge,
    \end{aligned}
\end{equation}
which yields:
    \begin{equation}
    \mathbf{A}_t= \begin{bmatrix}
    \mathbf{0}_{3,3} & \mathbf{0}_{3,3}  & \mathbf{0}_{3,3}& \mathbf{0}_{3,3} & \mathbf{0}_{3,3}\\
    {(\mathbf{g})_\times} & \mathbf{0}_{3,3} & \mathbf{0}_{3,3} & \mathbf{0}_{3,3} & \mathbf{0}_{3,3} \\
    \mathbf{0}_{3,3} & \mathbf{I}_{3} & \mathbf{0}_{3,3} & \mathbf{0}_{3,3} & \mathbf{0}_{3,3}\\
    \mathbf{0}_{3,3} & \mathbf{0}_{3,3}  & \mathbf{0}_{3,3}& \mathbf{0}_{3,3} & \mathbf{0}_{3,3}\\
    \mathbf{0}_{3,3} & \mathbf{0}_{3,3}  & \mathbf{0}_{3,3}& \mathbf{0}_{3,3} & \mathbf{0}_{3,3}
    \end{bmatrix}.
    \label{error_jacobian}
    \nonumber
\end{equation}
As a result, the state estimates can be propagated using (\ref{Dynamics}) and covariance matrix can be updated using the Riccati equation associated with~\eqref{linear_dif_error}:

\begin{equation}
    \frac{d}{dt}\bar{\mathbf{X}}_t=f_{u_t}(\bar{\mathbf{X}}_t), \quad \frac{d}{dt} \mathbf{P}_t= \mathbf{A}_t \mathbf{P}_t + \mathbf{P}_t \mathbf{A}_t^T + \Bar{\mathbf{Q}}_t,
\end{equation}
where $\Bar{\mathbf Q}_t$ is the process noise covariance defined as 
$
    \Bar{\mathbf{Q}}_t
    \triangleq \text{Cov}(\Bar{{ \mathbf{w}}}_t)
    = Ad_{{\Bar{\mathbf{X}}_t}} \text{Cov}(\mathbf{w}_t) Ad_{\Bar{\mathbf{X}}_t} \label{Q_cov}
$.

\subsection{Measurement update}
The nonlinear measurement model (\ref{kinematics measurement equation}) does not follow the right-invariant form, so we use a first-order approximation to find the innovation as:

\begin{equation}
    \begin{gathered}
    \mathbf{H}_t \boldsymbol{\zeta}_t + \text{h.o.t}(\boldsymbol{\zeta}_t)
    \triangleq
        {h}(\bar{\mathbf X}_t)-{h}(\mathbf X_t).  \label{measurement_error}
    \end{gathered}
\end{equation}
Since $\boldsymbol{\eta}_t \approx \bar{\mathbf{X}}_t {\mathbf{X}}_t^{-1}\approx \mathbf{I}+ \boldsymbol{\zeta}_t^{\wedge}$, we can derive the following relationships between the true and estimated states with $\boldsymbol{\zeta}_t$:
\begin{equation*}
    \begin{aligned}
   {\Rightarrow}~ &
   \mathbf R_t^T \approx \bar{ \mathbf R}_t^T (\mathbf{I} +(\boldsymbol{\zeta}_{ R_t})_\times), 
   ~
   \R_t^T \approx \bR_t^T (\mathbf{I} +(\boldsymbol{\zeta}_ {\Delta R _t})_\times)\\
   & \mathbf v_t 
   \approx 
   (\mathbf{I} -(\boldsymbol{\zeta}_{ R_t})_\times)(\bar{\mathbf v}_t -\boldsymbol{\zeta}_{ v_t}),  
   ~  
   \p_t \approx (\mathbf I- ( \boldsymbol{\zeta}_{\Delta R_t})_\times)(\overline{\p}_t -\boldsymbol{\zeta}_{\Delta p_t}).
    \end{aligned}
\end{equation*}
Now $\mathbf H_t$ can be computed by differentiating (\ref{measurement_error}) after dropping the nonlinear terms:

\begin{equation}
\begin{aligned}
       \mathbf H_t=& [\mathbf{0}_{3,3} ,
       ~\bR_t \bar{\mathbf R_t}^T ,~ \mathbf{0}_{3,3},~ \mathbf h_{4} ,~ (\bR_t \boldsymbol{\omega}_t)_\times ],\\
    \mathbf h_{4}=&
    -(\bR_t \bar{\mathbf R}_t^T \bar{\mathbf v}_t)_\times 
    - (\bR_t \boldsymbol{\omega}_t)_\times (\bp_t)_\times
      \\ 
       &+(\bp_t)_\times (\bR_t \boldsymbol{\omega}_t)_\times 
       +({h_F} (\boldsymbol{\alpha}_t))_\times (\bR_t \boldsymbol{\omega}_t)_\times.
\end{aligned}
\label{eq: H matrix FK}
\end{equation}

{For 3-D position based measurement model, the element $\mathbf{h}_4$ of the matrix $\mathbf H_t$ is slightly different from~\eqref{eq: H matrix FK}. The ${h_F} (\boldsymbol{\alpha}_t))$ needs to be replaced as $\mathbf{d}_t^M$:}

\begin{equation}
\begin{aligned}
       \mathbf H_t=& [\mathbf{0}_{3,3} ,
       ~\bR_t \bar{\mathbf R_t}^T ,~ \mathbf{0}_{3,3},~ \mathbf h_{4} ,~ (\bR_t \boldsymbol{\omega}_t)_\times ],\\
    \mathbf h_{4}=&-(\bR_t \bar{\mathbf R}_t^T \bar{\mathbf v}_t)_\times - (\bR_t \boldsymbol{\omega}_t)_\times (\bp_t)_\times\\
       &+(\bp_t)_\times (\bR_t \boldsymbol{\omega}_t)_\times 
       +(\mathbf{d}_t^M)_\times (\bR_t \boldsymbol{\omega}_t)_\times.
\end{aligned}
\end{equation}

Now we can express the update equation for our InEKF based on the InEKF methodology~\cite{barrau2016invariant}:

\begin{equation}
    \begin{gathered}
    \bar{\mathbf X}_t^+=\exp(\mathbf K_t (  \mathbf y_t-{h}(\bar{\mathbf X}_t ) ))\bar{\mathbf X}_t, \\
    \mathbf P_t^+=(\mathbf I-\mathbf K_t \mathbf H_t)\mathbf P_t^- (\mathbf I-\mathbf K_t \mathbf H_t)^T +\mathbf K_t \mathbf N_t \mathbf K_t^T,
    \end{gathered}
\end{equation}
where $\bar{\mathbf X}_t^+$ and $\mathbf P_t^+$ are the updated values.
Here, the Kalman gain $\mathbf K_t$ and measurement noise covariance $\mathbf N_t$ are defined as:
$
\mathbf K_t 
=\mathbf P_t \mathbf H_t ^T\mathbf S_t^{-1}$, 
$\mathbf S_t=\mathbf H_t \mathbf P_t^{-}\mathbf H_t^T+\mathbf N_t$,
and
$\mathbf N_t=\bar{\mathbf R}_t \bR_t^T \text{Cov}(\mathbf n_t) \bR_t \bar{\mathbf R}_t^T$.

\begin{algorithm}
    
    \label{alg:pseudo code}
    \caption{Proposed InEKF design}
    \begin{algorithmic}
    \State Initialize $\bar{\mathbf{X}} \in G$ and $\mathbf{P}=\mathbf{P}^T>0$

    \For{iteration=$1,2,\ldots$}
    
        \If{foot contact is detected}
            \State Propagation step:
            
            \State $\frac{d}{dt}\mathbf{\bar{X}}_t=f_{u_t}(\mathbf{\bar{X}}_t)$
            ,~$\frac{d}{dt} \mathbf{P}_t= \mathbf{A}_t \mathbf{P}_t + \mathbf{P}_t \mathbf{A}_t^T + \Bar{\mathbf{Q}}_t$
            
            \State Measurement update step:
            
            \State $\mathbf K_t =\mathbf P_t \mathbf H_t ^T\mathbf S_t^{-1}$,
            
            \State $\mathbf S_t=\mathbf H_t \mathbf P_t^{-}\mathbf H_t^T+\mathbf N_t$,
            \State $\mathbf N_t=\bar{\mathbf R}_t \bR_t^T \text{Cov}(\mathbf n_t) \bR_t \bar{\mathbf R}_t^T$
                \If {using forward kinematics based measurement}
                    \State $\mathbf y = - J(\tilde{\boldsymbol{\alpha}_t})  \dot{\tilde{\boldsymbol{\alpha}_t}}$
                \ElsIf{using 3-D vector based measurement}
                    \State $\mathbf y = - \tilde {\mathbf{v}}^{M}_t$
                \EndIf
            \State $\bar{\mathbf X}_t^+=\exp(\mathbf K_t (  \mathbf y_t-{h}(\bar{\mathbf X}_t^- ) ))\bar{\mathbf X}_t^-$
            \State $\mathbf P_t^+=(\mathbf I-\mathbf K_t \mathbf H_t)\mathbf P_t^- (\mathbf I-\mathbf K_t \mathbf H_t)^T +\mathbf K_t \mathbf N_t \mathbf K_t^T$
          \EndIf
        \EndFor
	\end{algorithmic} 
\end{algorithm}

\subsection{Observability analysis}
With new states and measurement model introduced, we need to analyze the observability of the whole system.
Here we only include the linear observability analysis, and leave the nonlinear observability analysis for future work. Similar to \cite{huang2010observability}, we analyze the observability around the operating point (i.e., the latest estimate that the system is linearized about).

The discrete filter is expressed as:
$
 \mathbf X_{k+1}= \boldsymbol{\Phi}_k \mathbf X_k$ and $\mathbf y_k=\mathbf H_k\mathbf X_k,
$
where $\Delta t=t_{k+1} -t_k$ is the duration of the propagation step with $t_k$ the timing of the $k^{th}$ update.
$\boldsymbol{\Phi}_k$ is the discrete-time state transition matrix, which can be computed as:

\begin{equation}
    \begin{gathered}
    \boldsymbol{\Phi}_k=\exp_m(\mathbf A_k\Delta t)=\begin{bmatrix}
     \mathbf I_{3} & \mathbf 0_{3,3}  & \mathbf 0_{3,3}& \mathbf 0_{3,3} & \mathbf 0_{3,3}\\
    (\mathbf g)_\times \Delta t & \mathbf I_{3} & \mathbf 0_{3,3} & \mathbf 0_{3,3} & \mathbf 0_{3,3} \\
    \frac{1}{2}(\mathbf g)_\times \Delta t^2 & \mathbf I_3 \Delta t & \mathbf I_{3} & \mathbf 0_{3,3} & \mathbf 0_{3,3}\\
    \mathbf 0_{3,3} & \mathbf 0_{3,3}  & \mathbf 0_{3,3}& \mathbf I_{3} & \mathbf \mathbf 0_{3,3}\\
    \mathbf 0_{3,3} & \mathbf 0_{3,3}  & \mathbf 0_{3,3}& \mathbf 0_{3,3} & \mathbf I_{3}
    \end{bmatrix}.
    \end{gathered}
\end{equation}

Then the observability matrix $\mathbf{O}$ can be computed as: 
\begin{equation*}
    \begin{gathered}
    \mathbf{O}=\begin{bmatrix}
     \mathbf H^-_k\\
      \mathbf H^-_{k+1} \boldsymbol{\Phi}_k^+\\
      \mathbf H^-_{k+2} \boldsymbol{\Phi}_{k+1}^+\boldsymbol{\Phi}_{k}^+\\
      \vdots \\
    \end{bmatrix}
    =\left[
   \begin{array}{c*{5}{@{\,}c}}
    \mathbf 0_{3,3} \ & \R_k^- \mathbf R_k^{-^{T}}  & \mathbf 0_{3,3} \ & o_{0,4} \  & (\R_k^- \boldsymbol{\omega}_k)_\times \\
    o_{1,1} & o_{1,2} & \mathbf 0_{3,3} & o_{1,4} & o_{1,5} \\
    \vdots & \vdots   & \vdots & \vdots& \vdots \\
    o_{4,1}& o_{4,2} & \mathbf 0_{3,3} & o_{4,4} & o_{4,5}\\
    \vdots & \vdots   & \vdots & \vdots& \vdots \\
     \end{array} 
   \right],
   \label{observability matrix}
    \end{gathered} 
\end{equation*}
where $(.)^+_k$ denotes the updated estimated state at time $t_k$, $(.)^-_k$ is the estimated state at time $t_k$ after the propagation step.
Here the terms $o_{i,1}$, $o_{i,2}$, $o_{i,4}$, and $o_{i,5}$ ($i \in \mathbb{N}^+$) are defined as:
\begin{equation*}
    \begin{aligned}
    o_{i,1}=&i (\R^-_{k+i}\mathbf R_{k+1}^{-^{T}} (\mathbf g)_\times\Delta t); 
    \quad
    o_{i,2}=\R^-_{k+i}\mathbf R_{k+1}^{-^{T}};
    \\
    o_{i,4}=&-(\R^-_{k+i}\mathbf R_{k+1}^{-^{T}} \mathbf v^-_{k+i})_\times- (\R^-_{k+i}\omega_{k+i})_\times (\p_{k+i}^-)_\times\\
    &+(\p_{k+i}^-)_\times (\R^-_{k+i} \boldsymbol{\omega}_{k+i})_\times 
    +( FK(\boldsymbol{\alpha}_{k+i}))_\times (\R^-_{k+i} \boldsymbol{\omega}_{k+i})_\times;\\
    o_{i,5}=&(\R^-_{k+i} \boldsymbol{\omega}_{k+1})_\times.
    \end{aligned}
\end{equation*}

To analyze the observability for each variable of interest, we need to see if the corresponding column vectors in $\mathbf O$ are linearly independent. From the observability matrix $\mathbf O$, it can be seen that the position of the IMU in the world frame is completely non-observable. The yaw angle of the IMU is also non-observable as the third column of the matrix $\mathbf g_\times$ is always zero. These results are similar to what was reported in \cite{hartley2020contact}. The observability of $\p$ and $\R$ depends on the rotational movement. For example, if the human is completely stationary ($\boldsymbol{\omega},\mathbf v= \mathbf 0$), both $\p$ and $\R$ are completely unobservable, as all corresponding columns in $\mathbf O$ are zero. If $\boldsymbol{\omega}=\mathbf 0, \mathbf v \neq0$, then $\p$ is completely non-observable, while it can be shown at least one of the components of $\R$ is observable. In general, if the direction of $\boldsymbol{\omega}$ is constant, then $\p$ will be non-observable in at least one direction.

\begin{figure}[t]
    \centering
    \includegraphics[width=0.45\textwidth]{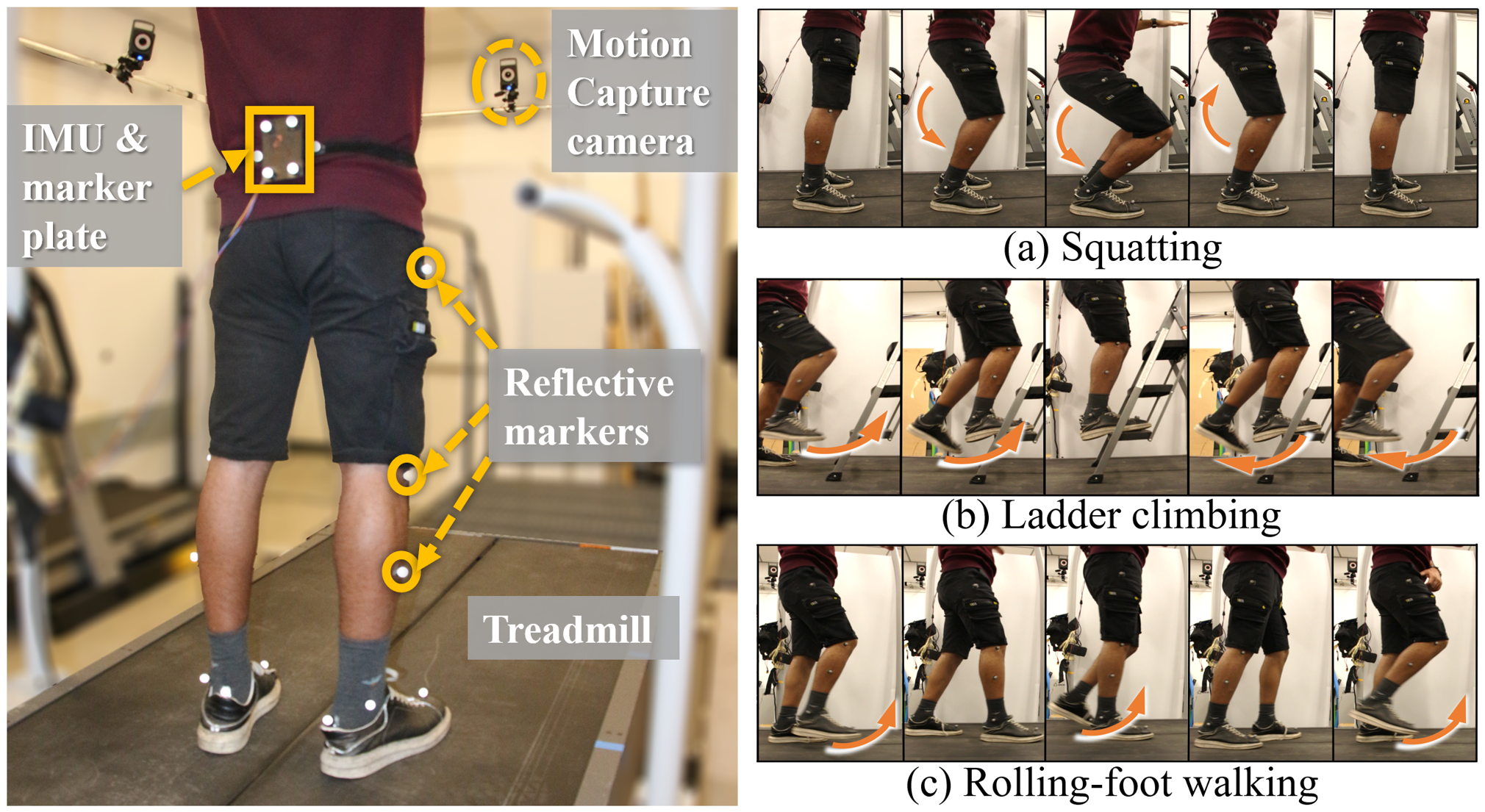}
    \vspace{-0.15 in}
    \caption{The left figure shows the experimental setup used to collect the ground truth of the estimated state variables and the sensor data needed by the proposed filter. The right figure shows time-lapse figures of the three motion types.
    The arrows in the subplots indicate the direction of the pelvis movement (in subplot (a)) and swing foot movement (subplots (b) and (c)).}
    \label{fig: exp_setup_time_lapse}
    \vspace{-0.2 in}
\end{figure}

\section{Experimental Results}
\label{sec: results}

\subsection{Experimental setup and protocol}

Experiments were performed in a motion capture laboratory at Arizona State University (ASU) with 3 participants (two males and one female, $27 \pm 3$ years old, $174\pm17 cm$, $77 \pm 17 kg$). The study was approved by institutional review board at ASU (STUDY00011266) and University of Massachusetts Lowell (20-057-YAN-XPD).

\noindent \textbf{Sensor setup.} Twelve IR motion capture cameras (Vicon, Oxford, UK) and sixteen reflective markers were used to build a lower-limb model for each subject via Vicon Nexus 2.8. Using this model, 3-D joint angles (hip, knee, and ankle) and pelvic position and orientation were estimated.
Four extra markers were attached to a plate that rigidly houses the IMU, which is used to build a rigid body model in Nexus to acquire the ground-truth pose of the IMU/body frame (as shown in the left figure of Fig.~\ref{fig: exp_setup_time_lapse}).
The IMU (BNO085, New York, NY) was placed on the back of the subject close to the pelvis (the placement of the IMU with regard to the pelvic center is shown in Fig. \ref{fig: coordinates represents}). The accelerometer and gyroscope data were recorded using a data acquisition board (Arduino UNO, Boston, MA). The data were later synchronised with the motion captured data using a trigger signal from the Vicon system. The experiments were done on an instrumented dual-belt treadmill equipped with force plates (Bertec Corp., Columbus, OH) that record ground contact forces.

\begin{table}[t]
\centering
\caption{Noise characteristics}

\label{table: noise characteristics}
\begin{tabular}{ |c||c|c|  }
 \hline   
 \multirow{2}{*}{Measurement type} & Noise SD
  & Noise SD\\ &  (proposed InEKF) &  (existing InEKF) \\
 \hline
 Linear acceleration & 0.2 m/s\textsuperscript{2} &  0.2 m/s\textsuperscript{2}\\ 
 Angular velocity & 0.05~rad/s  & 0.05 rad/s\\  
 Kinematics measurement & 0.5~m/s & 0.1 m\\
 Placement offset ($\p$, $\R$) & (0.05 m, 0.05 rad) & NA \\
 Contact velocity & NA & 0.05 m/s\\
 \hline
\end{tabular}

\end{table}

\noindent \textbf{Movement types.}
Each participant was asked to perform three types of motion: squatting, ladder climbing, and rolling-foot walking (see the right figure of Fig.~\ref{fig: exp_setup_time_lapse}). Two trials were performed for each motion type, each for 1.5 minute.

\subsection{Data Processing} 
\label{sec: data_Processing}

\noindent \textbf{Filters compared.}
The proposed filter is compared with a state-of-the-art InEKF~\cite{hartley2020contact}. The existing InEKF was originally designed for a Cassie series bipedal robot.
In the existing filter, the IMU and measurement frames are well aligned.
The state variables of the existing filter are the IMU orientation, velocity, and position and the contact foot position, all expressed in the world frame.
The kinematic measurements of the existing filter are the contact foot positions with respect to the measurement frame expressed in the measurement frame.
Unlike the proposed filter, the measurement model of the existing filter has an invariant observation form, which, in combination with the exponential form of the measurement update, renders the deterministic error update equation to be independent of state trajectories~\cite{barrau2016invariant}.
Also, the kinematic measurements of the existing filter have smaller noises since the Cassie series bipedal robot uses highly accurate leg encoders to formulate the forward kinematics chain. With a human subject, however, the IMU and measurement frames are not aligned, and the kinematics measurements have relatively large noises.

\noindent \textbf{Covariance settings.}
The noise characteristics for both filters are shown in Table \ref{table: noise characteristics}. The noise standard deviations of the linear acceleration and angular velocity are obtained from the IMU specifications provided by the manufacturer. To reach the better performance of both filters, these two noise standard deviations are slightly tuned around the nominal values. Note that covariance tuning is also reported in other InEKF designs \cite{brossard2020ai}.
It should also be noted that since the two filters use different measurement models ($\mathbf{d}^M$ is used as the measurement in \cite{hartley2020contact}), different noise covariance values are used for the two filters.
Moreover, the proposed filter considers the noise of the IMU angular velocity in the kinematics measurement noise term, 
while the same covariances are used for the common parameters of the two filters (linear acceleration and angular velocity).
During the tuning process it was observed that the performance of the proposed filter does not vary significantly within a relatively wide range of  parameters (e.g., the covariances matrices). Yet, the estimation performance degrades when the covariances are far from the optimal values.
The placement offset noises are only used in the proposed filter. Given that the placement offsets are relatively constant, the noise standard deviations of the placement offsets are set as small values.
The contact velocity noise is only accounted for in the existing filter, which is induced by contact foot slippage.

\noindent \textbf{Initial estimation errors.}
To demonstrate the accuracy and convergence rate of both filters under large estimation errors, a relatively wide range of initial estimation errors of the IMU/body velocity and orientation are used.
They are respectively chosen to be uniformly within $[-1,1]$ m/s and $[-20, 20]$ degrees across 50 trials.

\noindent \textbf{Filter performance indicators.}
To evaluate the filter performance, we choose to use three common indicators~\cite{barrau2016invariant,hartley2020contact,teng2021legged}: (a) computation time (for assessing the filter’s capability in real-time implementation); (b) convergence rate (for evaluating how rapidly the estimation error reaches the steady state); and (c) estimation accuracy (for testing the accuracy during transient and steady-state periods).

\subsection{Results}

\begin{figure}[t]
    \centering
    \includegraphics[width=0.43\textwidth]{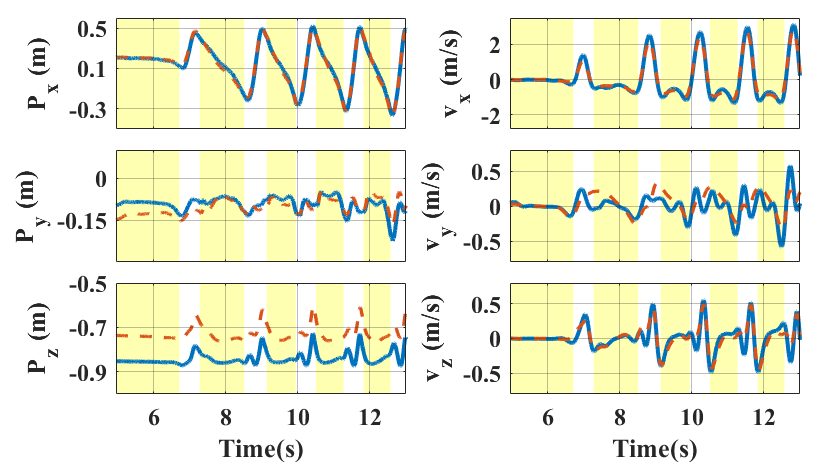}
   
    \caption{Relative position (left) and velocity (right) of the right toe in the measurement/pelvis frame during one of the rolling-foot walking trials, obtained by using joint angle readings and forward kinematics (blue, solid line) and by using the marker positions of the toe and measurement plate returned by motion capture system (red, dashed line).
    The yellow shaded areas indicate the periods during which the right foot contacts the ground.}
    \label{fig: FK measur}

\end{figure}

\begin{figure*}[ht]
    \centering 
    \includegraphics[width= 0.98\textwidth]{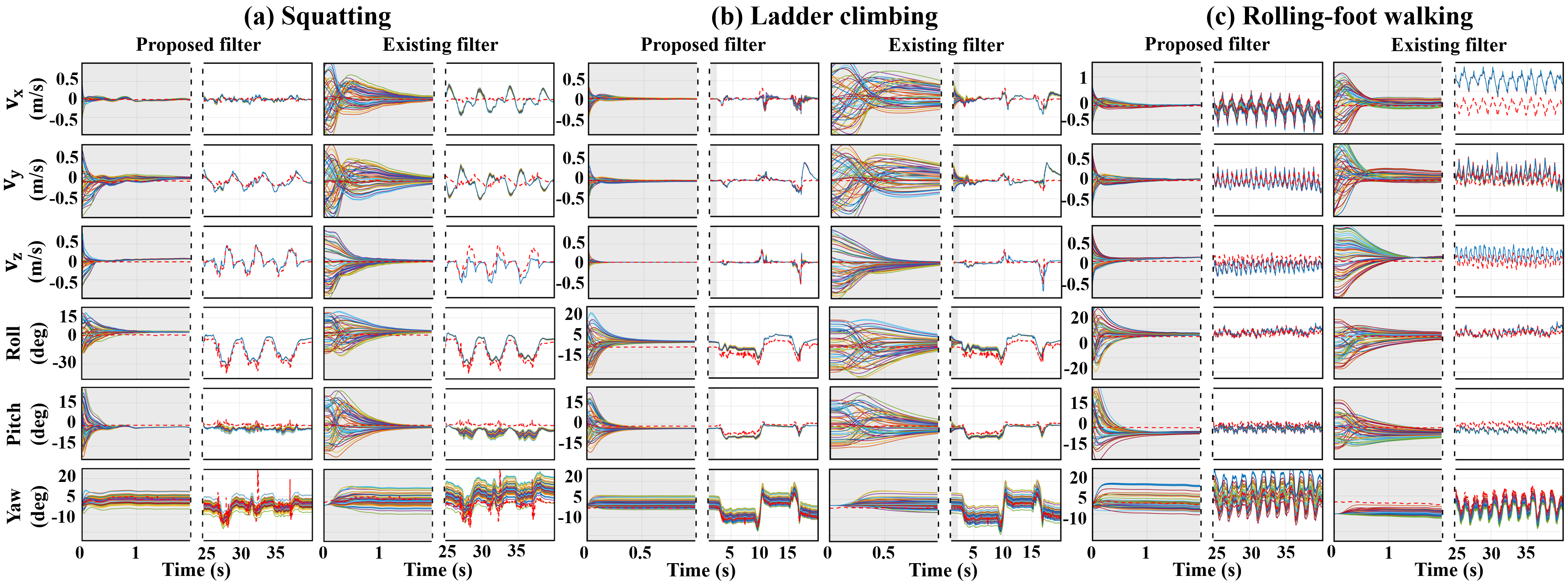}

    \caption{Estimation results of the velocity and orientation of the subject 2's body/IMU frame during (a) squatting motion, (b) ladder climbing, and (c) rolling-foot walking under the proposed and existing filters.
    The same sensor data set is used, including the raw data returned by the IMU at the trunk and the leg joint angle data provided by the motion caption system.
    The solid lines are the state estimates corresponding to different initial errors.
    The red, dashed lines are the ground truth.
    The gray and white backgrounds indicate the periods of initial and steady-state movement, respectively. The $x$-, $y$-, and $z$-directions are the lateral, forward and vertical directions, respectively.}
    \label{fig: all motion est}

\end{figure*}

To illustrate the accuracy of the forward kinematics based measurement~\eqref{kinematics measurement equation}, Fig.~\ref{fig: FK measur} compares it with its reference obtained from the motion capture system.
As noted in Sec.~\ref{alg:pseudo code}, the measurement associated with a given leg is fed into the update step of the filter only when the leg is in contact with the ground (highlighted by yellow shaded areas in Fig.~\ref{fig: FK measur}) .

Table \ref{table: RMSEs} displays the average root mean square error (RMSE) values of different motions and different kinematics measurements for all three subjects.
During the initial period, subjects are standing still, and the estimation algorithms just start.
During the steady-state motion period, subjects are doing continuous movements with different motion types.
The average RMSE values of the variable ``V'' are obtained from the estimated IMU velocity of all three axes for all subjects with the same motion type, while the average RMSE values of the variable ``O'' are obtained from the estimated IMU orientation of roll and pitch angles for all subjects with the same motion type.
The ``FK'' portion shows the RMSE values of the estimation results with the forward kinematics measurements in \eqref{kinematics measurement equation}, and the ``3-D vector'' part indicates the RMSE values of the estimation results under the 3-D vector kinematics measurements in \eqref{final kinematics measurement equation 3-D vector}.

Figure~\ref{fig: all motion est} shows the comparison of the IMU velocity and orientation estimation results of one subject with forward kinematics measurement under different motions and filters but the same sensor data set.
The grey shaded and white backgrounds indicate the initial period and steady-state motion period, respectively.

\begin{table}[t]
\centering
\caption{Average RMSE values for the three participants and motions. $V$ and $O$
refer to the body/IMU velocity and orientation.}

\adjustbox{width=0.48\textwidth, center}{
\begin{tabular}{ |>{\centering\arraybackslash}m{0.7cm}|
    >{\centering\arraybackslash}m{0.8cm}|
    >{\centering\arraybackslash}m{0.9cm}|
    >{\centering\arraybackslash}m{0.9cm}|
    >{\centering\arraybackslash}m{0.8cm}|
     >{\centering\arraybackslash}m{0.9cm}|
    >{\centering\arraybackslash}m{0.8cm}|} 
 \hline   
  \multirow{2}[3]{=}{\centering Time period} & \multirow{2}[2]{=}{\centering Motion type} &
  \multirow{2}[3]{=}{\centering Variable} & \multicolumn{2}{c|}{\centering FK} & \multicolumn{2}{c|}{\centering 3-D vector}\\
  \cline{4-7} 
  & & &  Proposed InEKF &  Existing InEKF &  Proposed InEKF &  Existing InEKF\\ 
 \hline
 \multirow{2}{=}{\centering Initial} 
    &  \multirow{2}{=}{\centering Stand} 
    & $V$ (m/s) & $0.062$ &  $0.226 $ & $0.063 $ &  $0.210 $   \\
    \cline{3-7}
    &  &  $O$ (deg) &  $4.484$  &  $6.616 $ &  $4.737 $  &  $6.643 $\\
 \hline
 \multirow{6}{=}{\centering Steady state} 
    & \multirow{2}{=}{\centering Squat} 
    &  $V$ (m/s) &  $0.067 $ &  $0.147 $ &  $0.041 $ &  $0.152 $ \\
    \cline{3-7}
    &  &  $O$ (deg) &  $4.745$  &  $4.909 $ &  $4.832 $ &  $4.727 $\\
    \cline{2-7}
    & \multirow{2}{=}{\centering Roll.-foot} 
    &  $V$ (m/s) &  $0.237$ &  $0.467 $ &  $0.103 $ &  $0.442 $\\
    \cline{3-7}
        &  &  $O$ (deg) &  $2.909$  &  $2.984 $ &  $2.685 $ &  $2.838 $\\
    \cline{2-7}
    & \multirow{2}{=}{\centering Ladder climb.} 
    &  $V$ (m/s) &  $0.072 $ &  $0.135 $ &  $0.047 $ &  $0.14 $ \\
    \cline{3-7}
    &  &  $O$ (deg) &  $3.13 $  &  $3.11 $ &  $3.12 $ &  $3.199 $\\
 \hline
\end{tabular}}
\label{table: RMSEs}

\end{table} 

\noindent \textbf{Computational time.}
MATLAB R2020b was used to process the experimental data sets with both filters.
The average computational time for one filter loop of the proposed filter is about 0.6 ms, and the average computational time for one filter loop of the existing filter is about 3 ms. 
Both are sufficiently fast for typical human movement monitoring.

\noindent \textbf{Convergence rate.}
By investigating the initial period of the estimation results figure (Fig.~\ref{fig: all motion est}) and the RMSEs table (Table~\ref{table: RMSEs}), it is obvious that the proposed filter converges faster than the existing filter, 
 driving the estimation error close to the ground truth within 0.6 s.

\noindent \textbf{Estimation accuracy.}
The results during the steady-state periods in both Fig.~\ref{fig: all motion est} and Table~\ref{table: RMSEs} indicate that under both filters the estimated roll and pitch angles of the IMU converge to a small neighborhood around their ground truth.
Yet, the yaw angle of the IMU is not observable under both filters.
Also, the overall accuracy of the IMU velocity estimation under the proposed filter is better than the existing filter.

\subsection{Discussion} \label{sec: discussion}
This subsection discusses the performance of the proposed InEKF with respect to the existing InEKF.  

\noindent \textbf{Forward kinematics vs. 3-D vector measurement.}
Incorporating forward kinematics measurement (i.e., \eqref{kinematics measurement equation}) introduces considerable uncertainties to the filtering system for human motion estimation (as depicted in Fig.~\ref{fig: FK measur}).
The uncertainties in the forward kinematics measurement could be induced by various sources, such as imperfect marker placement, shift of markers on skin or garment, inaccurate parameters (e.g., body segment lengths) and structure of human kinematics chain.
These factors lead to the less accurate estimation of both body velocity and orientation compared to directly using the 3-D vector measurement, for the proposed InEKF, as shown in Table~\ref{table: RMSEs}. 
Yet, the proposed filter with forward kinematics measurement has a better overall performance compared with the existing filter.
This highlights the importance of modeling the sensor placement offset in ensuring effective filtering under relatively less accurate forward kinematics measurement. 

\noindent \textbf{Convergence rate.}
As discussed in Sec.~\ref{sec: data_Processing}, the process and measurement covariance matrices were well tuned for both filters.
While the higher convergence rate can be a result of the explicit treatment of sensor placement offset in the proposed filter, additional human studies are needed to confirm that covariance matrices used in the two filters match the actual sensor characteristics and ensure a fair comparison.

\noindent \textbf{Steady-state estimation.}
From the average RMSE results in Table II, we can see that the proposed filter reaches a higher steady-state accuracy in the velocity estimation compared with the existing filter while they achieve similar accuracy in orientation estimation, given properly tuned noise covariance (and kinematics parameters for the forward kinematics based measurement model).
During the tuning process it was observed that both filters had better performance with relatively small angular velocity noise covariance. 
This indicates that orientation estimation was relied more on the process model for the steady-state period during which the large initial errors have already been corrected using the measurement model.
Therefore, both the proposed and existing InEKFs show an almost similar performance in orientation estimation during the steady-state period.
Nonetheless, velocity estimation is relatively more dependent on the measurements.
Therefore, the proposed filter has a superior performance in steady-state velocity estimation as it benefits from more accurate measurement updates thanks to its offset treatment.   

\noindent \textbf{Different motion estimation.}
From Table.~\ref{table: RMSEs}, it is evident that RMSE of the velocity estimations in rolling-foot walking is higher compared to the other two motion types.
Walking is a more dynamic task compared to the other two, making the estimation more challenging.
The static-foot assumption is more likely to be violated
\cite{gronqvist1995mechanisms} during foot-rolling.
Specifically, in rolling-foot walking, we observed relatively high errors in the $y$-direction of forward kinematic measurement.
When large errors are introduced in the measurement, it can also impact the offset estimation, weakening the advantage of the proposed InEKF with respect to the existing one.  
Since the existing filter does not consider the placement offsets, its velocity estimates have large final errors as shown in Fig.~\ref{fig: all motion est}.

\noindent \textbf{Limitations.}
One notable limitation of this work is the modeling of the IMU offset dynamics. It is assumed that the IMU offset has a slowly time-varying dynamics, which can be valid under some activities such as slow walking and stair climbing. However, under more aggressive movements such as running, the IMU might have large and sudden shifts relative to the body, which will violate this assumption. Other limitations include the practical difficulty in obtaining accurate joint angles for the forward kinematic model, and the validity of the assumption of the static foot-ground contact point (which was discussed through our results).

\section{Conclusion}

\label{sec: conclusion}

This paper introduced a right-invariant extended Kalman filter that explicitly considered the offsets between the IMU frame and the measurement frame.
The proposed filter design is an ``imperfect'' invariant extended Kalman filter since the process model satisfied the group affine property but the measurement model does not have the right-invariant observation form.
As demonstrated by experimental results among different subjects and motion types, the proposed filter has a low computational cost, and with properly tuned parameters (e.g., noise covariance and leg kinematics), it improves the convergence rate and estimation accuracy of the IMU velocity estimation compared with the existing filter.
This is largely because the IMU offset is treated as a noise source in the original filter while the proposed filter explicitly models and estimates it.
The observability analysis shows that the IMU positions and the rotation about the gravity vector were not observable whereas the IMU velocities and the rotations about the other two axes were observable.
The observability analysis matched with the experimental results.

In the future work, a more accurate forward kinematics model (e.g., obtained based on online estimation of limb lengths and other kinematics parameters) and explicitly treating IMU biases are needed to improve the filter performance.
Also, it is time consuming to tune the process and measurement covariances for different subjects and motion types.
A data-driven learning algorithm may be useful for solving this issue.
Finally, the forward kinematics measurement in this study assumes a fixed contact point on the foot, which may not be valid for movements involving a nonstationary contact point (e.g., during rolling-foot walking) and thus needs to be relaxed for a more realistic forward kinematics model.

\bibliographystyle{IEEEtran}
\bibliography{references.bib}

\begin{thebibliography}{10}
\providecommand{\url}[1]{#1}
\csname url@samestyle\endcsname
\providecommand{\newblock}{\relax}
\providecommand{\bibinfo}[2]{#2}
\providecommand{\BIBentrySTDinterwordspacing}{\spaceskip=0pt\relax}
\providecommand{\BIBentryALTinterwordstretchfactor}{4}
\providecommand{\BIBentryALTinterwordspacing}{\spaceskip=\fontdimen2\font plus
\BIBentryALTinterwordstretchfactor\fontdimen3\font minus
  \fontdimen4\font\relax}
\providecommand{\BIBforeignlanguage}[2]{{%
\expandafter\ifx\csname l@#1\endcsname\relax
\typeout{** WARNING: IEEEtran.bst: No hyphenation pattern has been}%
\typeout{** loaded for the language `#1'. Using the pattern for}%
\typeout{** the default language instead.}%
\else
\language=\csname l@#1\endcsname
\fi
#2}}
\providecommand{\BIBdecl}{\relax}
\BIBdecl

\bibitem{young2016state}
A.~J. Young and D.~P. Ferris, ``State of the art and future directions for
  lower limb robotic exoskeletons,'' \emph{IEEE Trans. Neur. Syst. Rehab.
  Eng.}, vol.~25, no.~2, pp. 171--182, 2016.

\bibitem{nordin2014assessment}
N.~Nordin, S.~Q. Xie, and B.~W{\"u}nsche, ``Assessment of movement quality in
  robot-assisted upper limb rehabilitation after stroke: a review,'' \emph{J.
  Neuroeng. Rehab.}, vol.~11, no.~1, pp. 1--23, 2014.

\bibitem{yumbla2021human}
E.~Q. Yumbla, Z.~Qiao, W.~Tao, and W.~Zhang, ``Human assistance and
  augmentation with wearable soft robotics: a literature review and
  perspectives,'' \emph{Cur. Rob. Rep.}, pp. 1--15, 2021.

\bibitem{chinimilli2019automatic}
P.~T. Chinimilli, Z.~Qiao, S.~M.~R. Sorkhabadi, V.~Jhawar, I.~H. Fong, and
  W.~Zhang, ``Automatic virtual impedance adaptation of a knee exoskeleton for
  personalized walking assistance,'' \emph{Rob. Auton. Syst.}, vol. 114, pp.
  66--76, 2019.

\bibitem{martinez2018velocity}
A.~Martinez, B.~Lawson, C.~Durrough, and M.~Goldfarb, ``A velocity-field-based
  controller for assisting leg movement during walking with a bilateral hip and
  knee lower limb exoskeleton,'' \emph{IEEE Trans. Rob.}, vol.~35, no.~2, pp.
  307--316, 2018.

\bibitem{luo2021reinforcement}
S.~Luo, G.~Androwis, S.~Adamovich, H.~Su, E.~Nunez, and X.~Zhou,
  ``Reinforcement learning and control of a lower extremity exoskeleton for
  squat assistance,'' \emph{Front. Rob. AI}, vol.~8, 2021.

\bibitem{del2014hybrid}
A.~J. Del-Ama, {\'A}.~Gil-Agudo, J.~L. Pons, and J.~C. Moreno, ``Hybrid
  {FES}-robot cooperative control of ambulatory gait rehabilitation
  exoskeleton,'' \emph{J. Neuroeng. Rehab.}, vol.~11, no.~1, pp. 1--15, 2014.

\bibitem{cheng2021real}
S.~Cheng, E.~Bol{\'\i}var-Nieto, and R.~D. Gregg, ``Real-time activity
  recognition with instantaneous characteristic features of thigh kinematics,''
  \emph{IEEE Trans. Neur. Syst. Rehab. Eng.}, vol.~29, pp. 1827--1837, 2021.

\bibitem{huang2021estimating}
R.~Huang, Z.~Peng, S.~Guo, K.~Shi, C.~Zou, J.~Qiu, and H.~Cheng, ``Estimating
  the center of mass of human-exoskeleton systems with physically coupled
  serial chain,'' in \emph{Proc. 2021 IEEE/RSJ Int. Conf. Intel. Rob. Syst.},
  pp. 9532--9539.

\bibitem{ojeda2007personal}
L.~Ojeda and J.~Borenstein, ``Personal dead-reckoning system for {GPS}-denied
  environments,'' in \emph{2007 IEEE Int. Workshop Safe., Secur. Resc. Rob.},
  2007, pp. 1--6.

\bibitem{yuan20133}
Q.~Yuan and I.-M. Chen, ``3-{D} localization of human based on an inertial
  capture system,'' \emph{IEEE Trans. Rob.}, vol.~29, no.~3, pp. 806--812,
  2013.

\bibitem{yuan2014localization}
Q.~Yuan and I.~Chen, ``Localization and velocity tracking of human via 3 {IMU}
  sensors,'' \emph{Sens. Act. A: Phys.}, vol. 212, pp. 25--33, 2014.

\bibitem{zhang2015whole}
Y.~Zhang, K.~Chen, J.~Yi, T.~Liu, and Q.~Pan, ``Whole-body pose estimation in
  human bicycle riding using a small set of wearable sensors,'' \emph{IEEE/ASME
  Trans. Mechatron.}, vol.~21, no.~1, pp. 163--174, 2015.

\bibitem{barrau2016invariant}
A.~Barrau and S.~Bonnabel, ``The invariant extended kalman filter as a stable
  observer,'' \emph{IEEE Trans. Autom. Contr.}, vol.~62, no.~4, pp. 1797--1812,
  2016.

\bibitem{hartley2020contact}
R.~Hartley, M.~Ghaffari, R.~M. Eustice, and J.~W. Grizzle, ``Contact-aided
  invariant extended kalman filtering for robot state estimation,'' \emph{Int.
  J. Rob. Res.}, vol.~39, no.~4, pp. 402--430, 2020.

\bibitem{gao2021invariant}
Y.~Gao, C.~Yuan, and Y.~Gu, ``Invariant extended {Kalman} filtering for hybrid
  models of bipedal robot walking,'' in \emph{Proc. of IFAC Mod., Est, and
  Contr. Conf.}, vol.~54, no.~20, 2021, pp. 290--297.

\bibitem{teng2021legged}
S.~Teng, M.~W. Mueller, and K.~Sreenath, ``Legged robot state estimation in
  slippery environments using invariant extended kalman filter with velocity
  update,'' in \emph{Proc. IEEE Int. Conf. Rob. Autom.}, 2021, pp. 3104--3110.

\bibitem{cheng2021asynchronous}
P.~Cheng, H.~Wang, V.~Stojanovic, S.~He, K.~Shi, X.~Luan, F.~Liu, and C.~Sun,
  ``Asynchronous fault detection observer for 2-d markov jump systems,''
  \emph{IEEE Trans. Cyb.}, 2021, in press.

\bibitem{zhu2021acc}
Z.~Zhu, M.~Rezayat, Y.~Gu, and W.~Zhang, ``Invariant extended kalman filtering
  for human motion estimation with imperfect sensor placement,'' in \emph{Proc.
  2022 Amer. Contr. Conf.}, accepted.

\bibitem{sola2018micro}
J.~Sola, J.~Deray, and D.~Atchuthan, ``A micro lie theory for state estimation
  in robotics,'' \emph{arXiv preprint arXiv:1812.01537}, 2018.

\bibitem{kainz2017reliability}
H.~Kainz, D.~Graham, J.~Edwards, H.~P. Walsh, S.~Maine, R.~N. Boyd, D.~G.
  Lloyd, L.~Modenese, and C.~P. Carty, ``Reliability of four models for
  clinical gait analysis,'' \emph{Gait \& posture}, vol.~54, pp. 325--331,
  2017.

\bibitem{huang2010observability}
G.~P. Huang, A.~I. Mourikis, and S.~I. Roumeliotis, ``Observability-based rules
  for designing consistent {EKF} {SLAM} estimators,'' \emph{Int. J. Rob. Res.},
  vol.~29, no.~5, pp. 502--528, 2010.

\bibitem{brossard2020ai}
M.~Brossard, A.~Barrau, and S.~abel, ``Ai-imu dead-reckoning,'' \emph{IEEE
  Trans. Intel. Veh.}, vol.~5, no.~4, pp. 585--595, 2020.

\bibitem{gronqvist1995mechanisms}
R.~Gr{\"o}nqvist, ``Mechanisms of friction and assessment of slip resistance of
  new and used footwear soles on contaminated floors,'' \emph{Erg.}, vol.~38,
  no.~2, pp. 224--241, 1995.

\end{thebibliography}

\begin{IEEEbiography}
[{\includegraphics[width=1in,height=1.25in,clip,keepaspectratio]{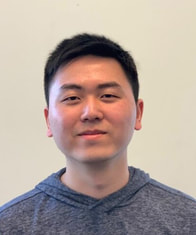}}]{Zenan Zhu}
received his B.S and M.S. degrees in Mechanical Engineering from UMass Lowell in 2018 and 2020, respectively. 
He is currently working toward a Ph.D. degree at UMass Lowell. His research interests include state estimation and optimization-based control of ankle exoskeletons for fatigue mitigation.
\end{IEEEbiography}

\begin{IEEEbiography}
[{\includegraphics[width=1in,height=1.25in,clip,keepaspectratio]{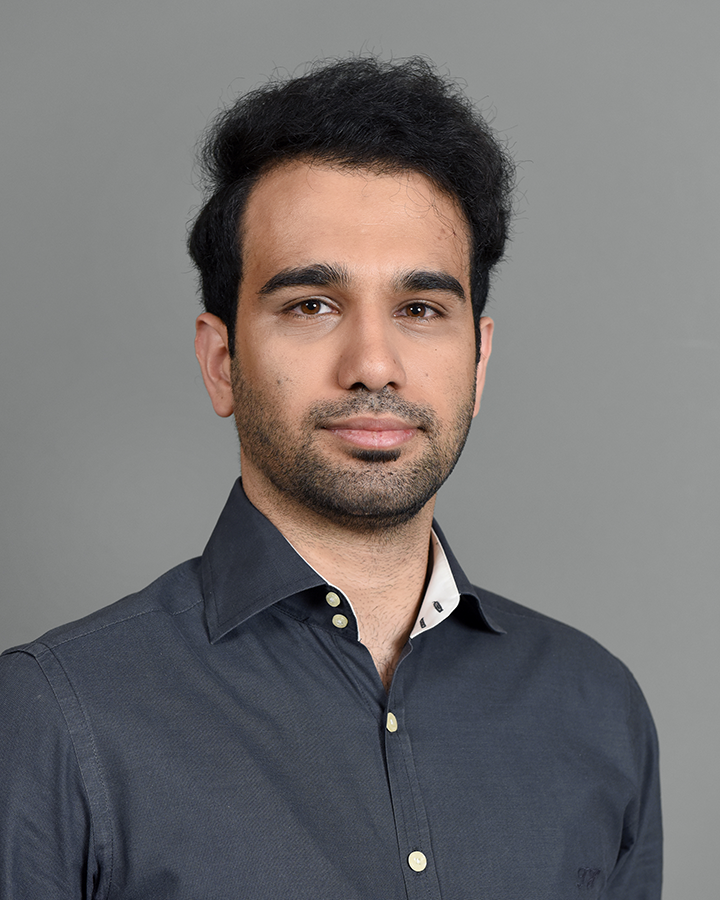}}]{Seyed Mostafa Rezayat Sorkhabadi}
received his M.S degree in Mechanical Engineering from Arizona State University in 2018. He is currently a Ph.D. candidate in the school for Engineering for Matter, Transport, and Energy at Arizona State University. His research interests include wearable sensors and robotics for gait rehabilitation and assistance.
   
\end{IEEEbiography}

\begin{IEEEbiography}
[{\includegraphics[width=1in,height=1.25in,clip,keepaspectratio]{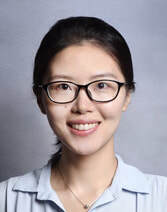}}]{Yan Gu}
received the B.S. degree in Mechanical Engineering from Zhejiang University, Hangzhou, China, in June 2011 and the Ph.D. degree in Mechanical Engineering from Purdue University, West Lafayette, IN, USA, in August 2017.
She joined the faculty of the School of Mechanical Engineering at Purdue University in 2022.
Prior to joining Purdue, she was an Assistant Professor with the Department of Mechanical Engineering at the University of Massachusetts Lowell.
Her research interests include nonlinear control, hybrid systems, legged locomotion, and wearable robots.
She was the recipient of the NSF CAREER Award in 2021.
\end{IEEEbiography}

\begin{IEEEbiography}
[{\includegraphics[width=1in,height=1.25in,clip,keepaspectratio]{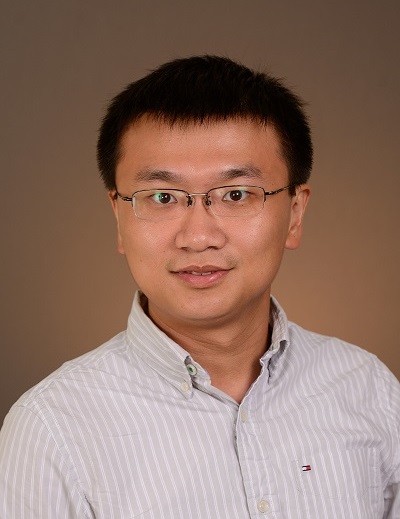}}]{Wenlong Zhang}
received the B.Eng. (Hons.) degree in control science and engineering from the Harbin Institute of Technology, Harbin, China, and the M.S. degree in mechanical engineering, the M.A. degree in statistics, and the Ph.D. degree in mechanical engineering from the University of California at Berkeley, Berkeley, CA, USA, in 2010, 2013, and 2015, respectively. He is currently an Associate Professor with the School of Manufacturing Systems and Networks at Arizona State University, where he directs the robotics and Intelligent Systems Laboratory (RISE Lab). His research interests include dynamic systems and control, human–machine collaboration, and autonomous aerial and ground vehicles.
\end{IEEEbiography}

\end{document}